\documentclass[letterpaper, 10 pt, conference]{ieeeconf}  

\IEEEoverridecommandlockouts                              

\usepackage{graphics} 
\usepackage{epsfig} 
\usepackage{mathptmx} 
\usepackage{times} 
\usepackage{amsmath} 
\usepackage{amssymb}  
\usepackage{bm}
\usepackage{url} 
\usepackage[nohyperlinks,printonlyused]{acronym} 
\usepackage[space]{cite}
\usepackage[bottom]{footmisc}

\newcommand{\specialcell}[2][c]{%
	\begin{tabular}[#1]{@{}c@{}}#2\end{tabular}}

\title{\LARGE \bf
Along Similar Lines: Local Obstacle Avoidance for Long-term Autonomous Path Following
}

\author{Jordy Sehn, Yuchen Wu, and Timothy D. Barfoot
\thanks{All authors are with the University of Toronto Institute for Aerospace Studies (UTIAS),
	 4925 Dufferin St, Ontario, Canada
	{\tt\small jordy.sehn@robotics.utias.utoronto.ca, yuchen.wu@robotics.utias.utoronto.ca, tim.barfoot@utoronto.ca}}%
}

\begin{document}
\addtolength{\textfloatsep}{-16pt} 
\addtolength{\abovecaptionskip}{-8pt} 

\maketitle
\thispagestyle{empty}
\pagestyle{empty}

\begin{abstract}
Visual Teach and Repeat 3 (VT\&R3), a generalization of stereo VT\&R, achieves long-term autonomous path-following using topometric mapping and localization from a single rich sensor stream. In this paper, we improve the capabilities of a LiDAR implementation of VT\&R3 to reliably detect and avoid obstacles in changing environments. Our architecture simplifies the obstacle-perception problem to that of place-dependent change detection. We then extend the behaviour of generic sample-based motion planners to better suit the teach-and-repeat problem structure by introducing a new edge-cost metric paired with a curvilinear planning space. The resulting planner generates naturally smooth paths that  avoid local obstacles while minimizing lateral path deviation to best exploit prior terrain knowledge. While we use the method with VT\&R, it can be generalized to suit arbitrary path-following applications. Experimental results from online run-time analysis, unit testing, and qualitative experiments on a differential drive robot show the promise of the technique for reliable long-term autonomous operation in complex unstructured environments.

\end{abstract}

\section{INTRODUCTION}

Robot navigation in unstructured outdoor environments presents a challenging-yet-critical task for many mobile robotics applications including transportation, mining, and forestry. In particular, robust localization in the presence of both short- and long-term scene variations without reliance on a global positioning system (GPS) becomes very difficult. Furthermore, the off-road terrain-assessment problem is non-trivial to generalize as the variety of potential obstacles increases, all of which require careful identification, planning, and control to prevent collisions.

Visual Teach and Repeat (VT\&R) \cite{Furgale2010} attempts to tackle these problems by suggesting that often it is sufficient for a mobile robot to operate on a network of paths previously taught by a human operator. During a learning phase (the \textit{teach pass}) the robot is manually piloted along a route whilst building a visual map of the environment using a single rich sensor such as a stereo camera. In the autonomous traversal phase (the \textit{repeat pass}), live stereo images are used to localize to the map with high precision and resiliency to lighting and seasonal changes \cite{Paton2016, Gridseth2021}.

\begin{figure}[thpb]
	\centering
	\includegraphics[scale=0.38]{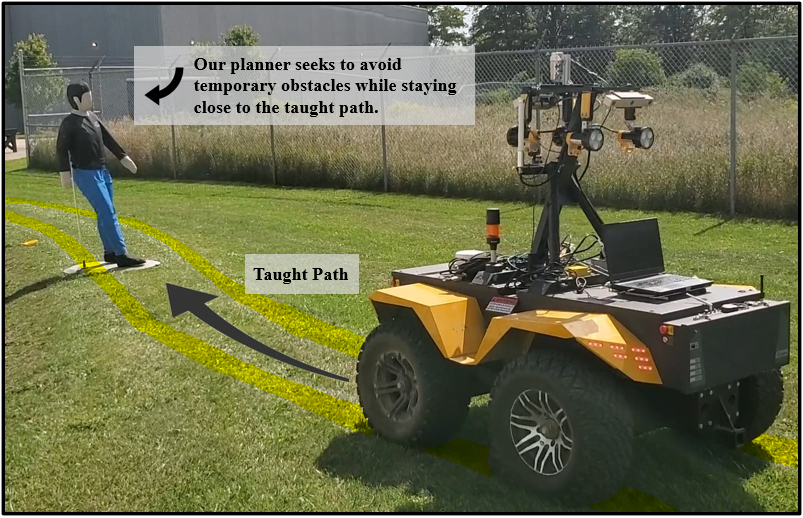}
	\caption{We present an architecture for locally avoiding unmapped obstacles along a reference path by extending sample-based motion planners to encourage trajectories with characteristics that best exploit prior path knowledge. Our system is validated on a Grizzly robot in real-world scenarios.}
	\setlength{\belowcaptionskip}{-12pt}
	\label{figure1}
\end{figure}
\setlength{\belowcaptionskip}{-12pt}

In practice, this architecture works well to address both the localization and terrain-assessment problems. By having a human operator drive the teach pass, we exploit the strong prior that the original path was traversable. It follows that, at least in the short term, it is very likely the taught path remains collision free and by following the path closely in the repeat, minimal terrain assessment is required.

With the advent of LiDAR implementations of teach and repeat \cite{McManus2012,Krusi2014,Cherubini2013} whose localization is less sensitive to viewpoint changes when repeating a path than stereo, we explore the possibility of temporarily deviating from the teach path to avoid new obstacles and increase the practicality of VT\&R. Specifically, we tailor our perception, planning, and control methods to exploit the teach-and-repeat problem structure while continuing to leverage the human terrain assessment prior whenever possible.


The primary contribution of our system is in the planning domain where we present a novel edge-cost metric that, when combined with a curvilinear planning space, extends the capabilities of a generic sample-based optimal motion planner to generate paths that naturally avoid local obstacles along the teach pass. In obstacle-free environments, the solution must stay on the original teach path and not cut corners. We then follow the output paths using Model Predictive Control (MPC) to enforce kinematic constraint and generate smooth trajectories. Critically, our edge metric encourages path solutions that strike a balance between minimizing both path length and lateral path deviation without using waypoints. Furthermore, we show that our new metric maintains the properties of the underlying planning algorithm by using an admissible sampling heuristic. 

We illustrate the effectiveness of our planner in simulation on a set of realistic path-following problems. The entire system is then integrated with VT\&R3 and we demonstrate the obstacle-avoidance capabilities on an unmanned ground vehicle (UGV) over several kilometers of long-term autonomous navigation in a variety of unstructured environments.

\section{RELATED WORK}

Obstacle avoidance in unstructured environments is a broad and well-studied field. Generally, the literature can be categorized into the solution of two sub-problems: (i) obstacle detection and (ii) local trajectory planning given a set of obstacles.

\subsection{LiDAR Obstacle Detection}
Obstacle detection in this work can be related to the problem of unsupervised object discovery from LiDAR change detection, which is often handled jointly with long-term mapping \cite{Ambrus2014,Finman2013,Ambrus2015,Fehr2017}. We are concerned with detecting changes in the environment from two sets of LiDAR measurements
or their derived representations (i.e., point clouds, voxel grids), with one considered a query and the other as a reference. The most direct approach is accumulating measurements from both sets into point clouds, computing the nearest-neighbour distances between points from the two clouds, and classifying changes by thresholding the distance \cite{Girardeau2005}. Although more sophisticated classification techniques have been explored in \cite{Andreasson2007,Nunez2009,Vieira2012}, decent results can be achieved if the point clouds are dense and there is no need to reason about occlusions.

This research is inspired by previous terrain-assessment works for VT\&R, which also rely on
change detection \cite{Berczi2016, Berczi2017}. In \cite{Berczi2016}, the terrain ahead of the robot is organized into robot-sized patches, each of which is a 2D grid storing the estimated terrain height. These patches are compared with those from previous experiences, and any significant difference causes the respective terrain to be marked untraversable. The approach is improved in \cite{Berczi2017}, with each cell storing a 1D Gaussian Mixture Model (GMM) with two Gaussians in the vertical direction. One Gaussian models the terrain height and roughness, and the other addresses noises due to overhanging vegetation. 

Our method learns from \cite{Berczi2017} to explicitly account for surface roughness in change detection using a Gaussian model, improving the detection of small obstacles in structured environments and applies the notion to LiDAR point clouds.

\subsection{Local Trajectory Planning}
In the teach-and-repeat problem structure, we have a relative reference path to follow and wish to temporarily deviate from the path to avoid obstacles. A simple way to achieve this was demonstrated by Krusi et al. \cite{Krusi2014}. They avoid local obstacles by generating a tree of trajectories at each time step and select the best one with respect to a cost function. While simple and effective, the output trajectories are not optimized for path-following, even in obstacle-free environments, an undesirable property for high-precision teach-and-repeat applications. 

A similar trajectory-sampling approach is used by Said et al. \cite{Said2021}. However, in this case the trajectories are sampled from a curvilinear coordinate space with axes defined by the longitudinal and lateral distances along the reference path, respectively. Using this method ensures that at least one of the sampled trajectories lies exactly on the reference path ensuring an optimal solution in the event there are no obstacles. Our method uses a very similar curvilinear planning space to exploit the same property within a sample-based planner, but differs greatly in the application. 

With the prevalence of many high-speed optimization solvers, direct path-tracking MPC has become a popular choice \cite{Faulwasser2015,Ji2017} for the ability to easily enforce system constraints and tune the behaviour of the controller. In these implementations, a quadratic cost function typically models the error between a robot's predicted state and a path given a sequence of control inputs to be optimized.

To avoid obstacles, Dong et al. \cite{Dong2020} adds additional state constraints to MPC. Normally obstacle constraints make the problem non-convex due to the spawning of additional path homotopy classes resulting in local minima. They are able to avoid this problem by approximating obstacles as low-dimensional linear convex hulls to find globally optimal solutions for simple vehicle passing manoeuvres. Lindqvist et al. \cite{Lindqvist2020} shows how the use of a nonlinear, non-convex solver such as Proximal Averaged Newton for Optimal Control (PANOC) \cite{Sathya2018} can successfully solve obstacle-avoidance problems by describing sets of obstacles as nonlinear inequality constraints and then relaxing them with penalty functions. These types of approaches work well in structured environments with few obstacles, but fail to scale well when the number and variety of obstacles increases.

Instead of trying to approximately solve non-convex MPC problems directly, an alternative strategy is to first run a separate path-planning algorithm to find a route that avoids obstacles and then reconnects with the global path. Liniger et al. \cite{Liniger2015} applies this idea to autonomous racing, using dynamic programming to find rough shortest-distance paths avoiding other vehicles. This path solution is representive of the most promising path homotopy class. By defining convex lateral state constraints on the MPC based on this initial homotopy solution, they are able to generate globally optimal trajectories at high speeds. Patterson et al. \cite{Patterson2021} uses an even simpler method that defines the lateral corridor of MPC constraints using geometric rules based on the location of obstacles relative to the road centre-line. 

Zhou et al. \cite{Zhou2022} run a variation of the sample-based planner Informed RRT* \cite{Gammell2014} to generate shortest-distance paths avoiding obstacles, and opt to track this path directly with MPC under the assumption that it will be collision free. Of these techniques, our method is most similar to \cite{Zhou2022}. The distinction is that the planner we use adopts our unique edge metric, which finds paths best exploiting our prior terrain knowledge by reducing lateral error with the reference and is suitable for large networks of paths.

\section{METHODOLOGY}

This section describes our approach for autonomous relative path following given a previously taught route. An overview of our architecture is shown in Fig. 2, of which each component is described in further detail.

\begin{figure}[t]
	\centering
	\includegraphics[scale=0.32]{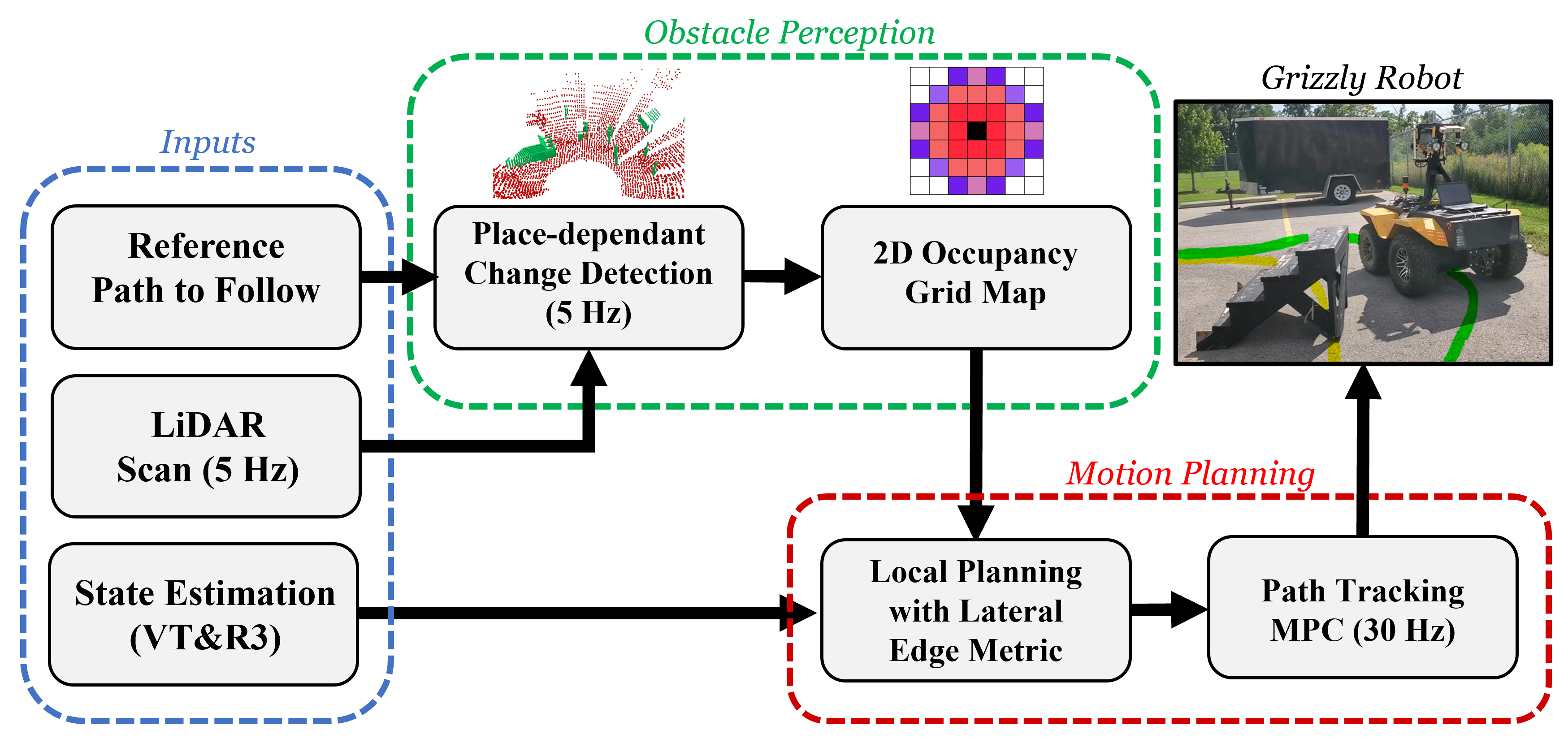}
	\caption{An overview of the proposed obstacle-avoidance system. A change-detection LiDAR perception module identifies previously unmapped structures and updates a locally planar 2D occupancy grid map. The planner finds paths that avoid obstacles using our laterally weighted edge-cost metric and a tracking MPC enforces kinematic constraints on the final trajectory.}
	\label{figure2}
\end{figure}

\subsection{Obstacle Detection}
As the obstacle-detection system is not the focus of evaluation for this work, we direct the reader to Wu \cite{Wu2022} for implementation details. At a high level, we treat obstacle detection as a change-detection problem between the environment's long-term static structures and the current LiDAR observation. If a point from the live scan fails to coincide with a mapped structure, it is likely from a new obstacle. 

We then adopt classical methods for classifying points in the scan based on a thresholding method similar to \cite{Girardeau2005}, while modelling surface roughness with a Gaussian in our classification metric as in \cite{Berczi2016}. Going forward we take for granted that all obstacles are reliably detected and projected onto local 2D occupancy grids for collision checking.




\subsection{Sample-based Planning Preliminaries and Notation}
While the selection of path planner is arbitrary for the application of our extensions, we employ Batch Informed Trees (BIT*) \cite{Gammell2015} as our baseline planner. BIT* is probabilistically complete, asymptotically optimal, and can be adapted to re-plan or \textit{rewire} itself in the presence of moving obstacles, making it an ideal candidate for our current and future applications.

BIT* performs as follows. A Random Geometric Graph (RGG) with \textit{implicit} edges is defined by uniformly sampling the free space around a start and goal position. An \textit{explicit} tree is constructed from the starting point to the goal using a heuristically guided search through the set of samples. Given a starting state $\mathrm{\mathbf{x}}_\mathrm{start}$ and goal state $\mathrm{\mathbf{x}}_\mathrm{goal}$, the function $\hat{f}(\mathrm{\mathbf{x}})$ represents an admissible estimate (i.e., a lower bound) for the cost of the path from $\mathrm{\mathbf{x}}_\mathrm{start}$ to $\mathrm{\mathbf{x}}_\mathrm{goal}$, constrained through $\mathrm{\mathbf{x}} \in \mathrm{X}$. 

Admissible estimates of the cost-to-come and cost-to-go to a state $\mathrm{\mathbf{x}} \in \mathrm{X}$ are given by $\hat{g}(\mathrm{\mathbf{x}})$ and $\hat{h}(\mathrm{\mathbf{x}})$, respectively, such that $\hat{f}(\mathrm{\mathbf{x}})$ = $\hat{g}(\mathrm{\mathbf{x}})$ + $\hat{h}(\mathrm{\mathbf{x}})$. Similarly, an admissible estimate for the cost of creating an edge between states $\mathrm{\mathbf{x}}, \mathrm{\mathbf{y}} \in \mathrm{X}$ is given by $\hat{c}(\mathrm{\mathbf{x}}, \mathrm{\mathbf{y}})$. Together, BIT* uses these heuristics to process and filter the samples in the queue based on their ability to improve the current path solution. The tree only stores collision-free edges and continues to expand until either a solution is found or the samples are depleted.

A new batch begins by adding more samples to construct a denser RGG. Had a valid solution been found in the previous batch, the samples added are limited to the subproblem that could contain a better solution.  Given an initial solution cost, $c_{\mathrm{best}}$, we can define a subset of states, $X_{\hat{f}} := \{\mathrm{\mathbf{x}} \in \mathrm{X} \Big| \hat{f}(\mathrm{\mathbf{x}}) \leq c_{\mathrm{best}} \}$ that have the possibility of improving the solution. When the metric for the edge-cost computation is Euclidean distance, the region defined by $\hat{f}(\mathrm{\mathbf{x}}) \leq c_{\mathrm{best}}$ is that of a hyperellipsoid with transverse diameter $c_{\mathrm{best}}$, conjugate diameter $\sqrt{c_{\mathrm{best}^2} + c_{\mathrm{min}^2}}$, and focii at $\mathrm{\mathbf{x}}_\mathrm{start}$ and $\mathrm{\mathbf{x}}_\mathrm{goal}$ \cite{Gammell2014}. 

The original implementation of BIT* is designed for shortest-distance point-to-point planning and is not customized for path following. In the remaining sections, we describe two modifications to adapt a generic sample-based  planner for the problem structure outlined in Section I.

\subsection{Curvilinear Coordinates}

Our first extension adds natural path-following by using an orthogonal curvilinear planning domain \cite{Barfoot2004}. A reference path is composed of a set of discrete 3 Degree of Freedom (DOF) poses $P = \{\mathbf{x}_{\mathrm{start}}, \, \mathbf{x}_1,\, \mathbf{x}_2, \hdots \, ,\, \mathbf{x}_{\mathrm{goal}}\}$ with $\mathbf{x} = (x, y, \psi)$ describing the Euclidean position and yaw. We define a curvilinear coordinate, $(p,q)$, representation of the path such that the $p$-axis, $p \in [0, p_\mathrm{len}]$ describes the longitudinal distance along the path, and the $q$-axis, $q \in [q_\mathrm{min}, q_\mathrm{max}]$ is the lateral distance perpendicular to each point $p$ on the path. $q_\mathrm{min}$ and $q_\mathrm{max}$  describe the lateral place-dependant bounds of the curvilinear space at each segment of the path.

\begin{figure}[b]
	\centering
	\includegraphics[scale=0.75]{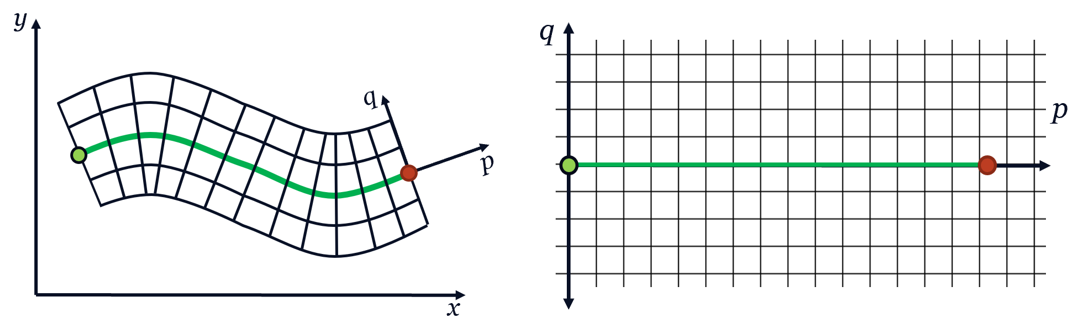}
	\caption{Left: A reference path in Euclidean coordinates shown in green, with the longitudinal and lateral components extended in a grid. Right: The corresponding representation of the path in curvilinear coordinates.}
	\label{figure3}
\end{figure}

A change in distance between subsequent poses, $\Delta p$, is computed as
\begin{equation} \label{eq:one}
	\Delta p = \sqrt{\Delta x^2 + \Delta y^2 + a \Delta \psi^2}.
\end{equation}

\noindent An aggregated $p$ value is stored for each discrete pose in a pre-processing step up to the total length of the path, $p_\mathrm{len}$. It is important to note that as part of \eqref{eq:one}, we incorporate a small term for changes in yaw along the repeat path tuned by a constant parameter $a$. This allows us to avoid singularities in the curvilinear coordinate space in the event of rotations on the spot by distinguishing between poses with identical positions but changing orientations.

A key observation from this definition is that all paths in Euclidean space become straight lines in $(p,q)$ space. By storing the $p$ values associated with each Euclidean pose from the reference path, we can uniquely map an arbitrary curvilinear point $(p_i,q_i)$ to its corresponding Euclidean point $(x_i, y_i)$ by interpolating to find a pose on the reference path closest to the target point and applying some basic trigonometry. 

While a unique map always exists from $(p,q)$ to Euclidean space, generally the reverse can not be guaranteed due to singularities. This proves to be problematic when considering the collision checking of obstacles. Our solution is intuitive: we run BIT* in $(p,q)$ space as normal, and perform all collision checks in Euclidean space by discretizing edges and mapping the individual points back to Euclidean space for query with the obstacle costmaps.





\subsection{Weighted Euclidean Edge Metric}


Consider the 2D Euclidean planning problem where the teach path is composed of a path connecting $\mathbf{x}_\mathrm{start}$ to $\mathbf{x}_\mathrm{goal}$. The usual cost of an edge connecting two arbitrary points in space $(x_1,y_1)$ to $(x_2, y_2)$ can be expressed generally as the length $c_{21}$:

\begin{equation} \label{eq:five}
	c_{21} = \int_{x_1}^{x_2} \sqrt{1 + \Big(\frac{dy}{dx}\Big)^2} dx.
\end{equation}

\noindent For our work, we incorporate an additional coefficient such that the cost of an edge increases as the lateral $y$ deviation over the length of the edge grows, scaled by a tuning parameter $\alpha$:
\begin{equation} \label{eq:seven}
	c_{21} = \int_{x_1}^{x_2}(1+ \alpha y^2) \sqrt{1 + \Big(\frac{dy}{dx}\Big)^2} dx.
\end{equation}

\noindent For a straight line, this integral becomes
\begin{align} \label{eq:eight}
	&c_{21} = \Bigg(1 + \frac{\alpha(y_2^3 - y_1^{3})}{3(y_2 - y_1)}\Bigg) \sqrt{(x_2 - x_1)^2 + (y_2 - y_1)^2}.
\end{align}

\noindent As $\Delta y$ approaches zero (horizontal edges) we have
\begin{align} \label{eq:nine}
	\begin{aligned}
		&\lim\limits_{\Delta y \rightarrow 0}
		\Bigg(1 + \frac{\alpha(y_2^3 - y_1^{3})}{3(y_2 - y_1)}\Bigg) \sqrt{(x_2 - x_1)^2 + (y_2 - y_1)^2} \\
		& \qquad \qquad \qquad \qquad \qquad \qquad \quad \,= (1 + \alpha y^2) |x_2 - x_1|,
	\end{aligned}
\end{align}

\noindent where $y_1 = y_2 = y$.

In \eqref{eq:eight} we obtain the Euclidean distance metric scaled by a coefficient to apply a penalty for lateral path deviation. While this edge metric works in this simple example of a straight-line path, the result is difficult to generalize when considering arbitrarily complex reference paths in Euclidean space. In curvilinear space, however, all reference paths become horizontal lines on the $p$-axis, allowing us to directly apply this idea for the edge-cost metric in BIT* by letting $y_i \rightarrow q_i$ and $x_i \rightarrow p_i$, respectively, in \eqref{eq:eight}.

Before using this new metric in BIT*, we must first evaluate the influence of the edge-cost to the informed sampling region that constrains the RGG sub-problem following an initial solution. Consider the estimated total cost, $\hat{f}(\mathrm{\mathbf{x}}) = \hat{g}(\mathrm{\mathbf{x}}) + \hat{h}(\mathrm{\mathbf{x}})$, to incorporate an arbitrary sample, $\mathbf{x} = (p,q)$, into the path solution in curvilinear coordinates as in Fig. 4.

\begin{figure}[t]
	\centering
	\includegraphics[scale=0.4]{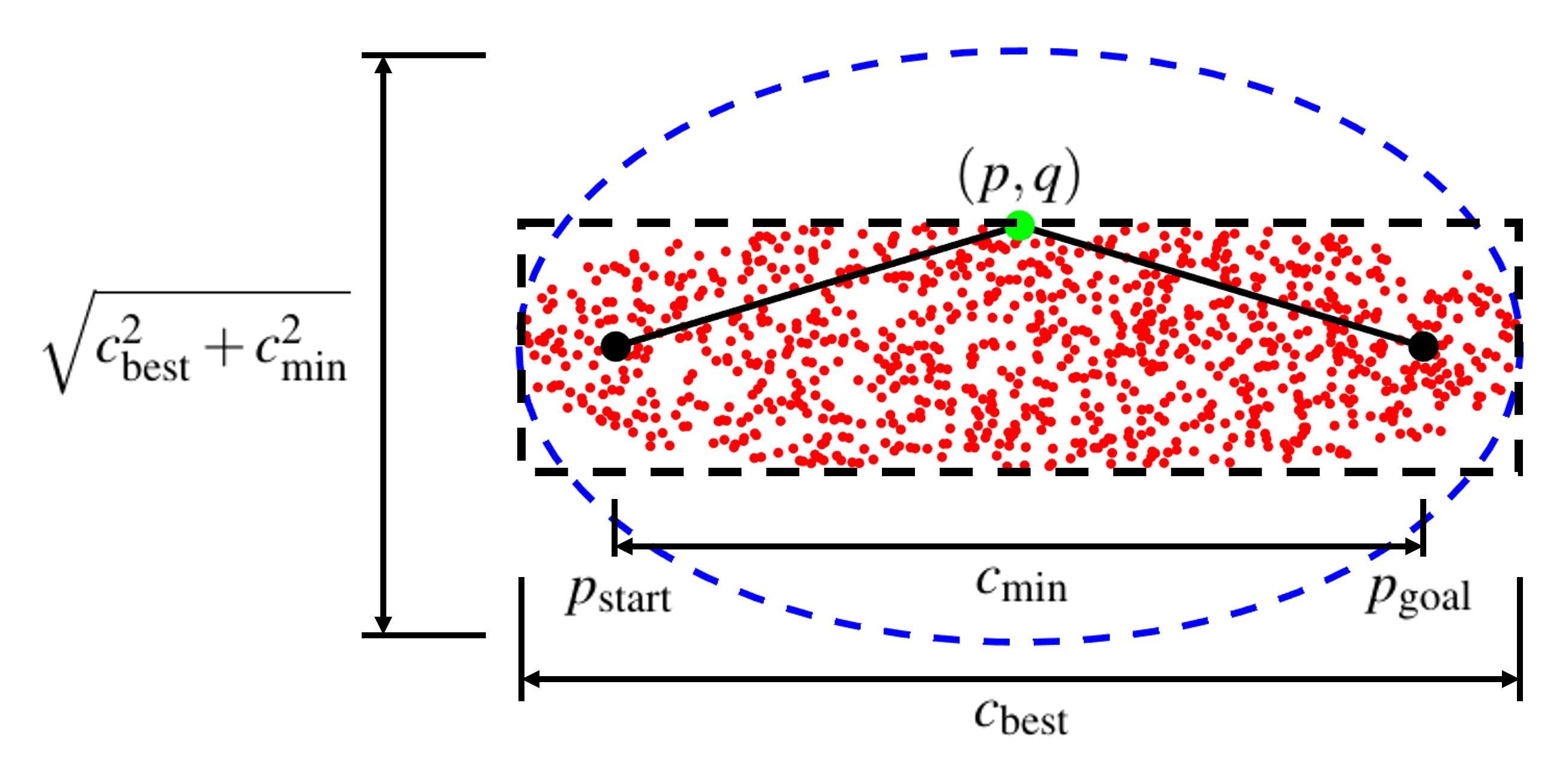}
	\caption{The informed sampling domain, $X_{\hat{f}}$, for the Euclidean distance edge metric (blue ellipse), and the conservatively bounded laterally weighted edge metric (black box) for $\alpha = 0.5$. Samples shown in red, were populated using rejection sampling and illustrate the true `eye-ball' distribution of the informed sampling region. As $\alpha$ tends to zero, the domains coincide.}
	\label{figure4}
\end{figure}

The cost is
\begin{align} \label{eq:ten}
	\scalebox{0.94}{
	$\begin{aligned} [t]
		\hat{f}(\mathrm{\mathbf{x}}) =& \Big(1 + \frac{\alpha}{3} q^2\Big) 
		\Big(\sqrt{(p - p_\mathrm{start})^2 + q^2} + 	\sqrt{(p - p_\mathrm{goal})^2 + q^2} \,\Big).
	\end{aligned}$
	}
\end{align}


\noindent If $\mathbf{x}$ is to improve the quality of a current solution cost, $c_\mathrm{best}$,
we require that $\hat{f}(\mathrm{\mathbf{x}}) \leq c_\mathrm{best}$. Rearranging the inequality, we have
\begin{align} \label{eq:eleven}
	\begin{aligned}
		&\Big(\sqrt{(p - p_\mathrm{start})^2 + q^2} + 	\sqrt{(p - p_\mathrm{goal})^2 + q^2} \,\Big) \\
		& \qquad \qquad \qquad \qquad \qquad \qquad \quad \leq \frac{c_\mathrm{best}}{\Big(1 + \frac{\alpha}{3} q^2\Big)} \leq c_\mathrm{best}.
	\end{aligned}
\end{align}

\noindent We note that the lateral scaling factor is always $\geq 1$ for all $q$ and that once again the left-hand term is simply the Euclidean distance edge metric. This result implies that, conservatively, we could sample from within the informed ellipse defined by the usual Euclidean edge cost (no lateral penalty) and the probabilistic guarantees will remain satisfied. 

On the other hand, \eqref{eq:eleven} indicates that our true informed sampling region exists as a denser subset within the original ellipse. While difficult to describe geometrically, we can visualize this region by randomly sampling from within the outer ellipse and rejecting points with heuristic costs larger than the bounds. As shown in red in Fig. 4, we see the true informed sampling region takes the shape of an `eye-ball' and for this problem has significantly smaller volume than the Euclidean distance ellipse. 


In practice, direct sampling from the eye-ball region is difficult and rejection sampling can be inefficient. However, it is possible to calculate the height of a conservative rectangle to bound the true informed sampling region and perform direct sampling from the bounding box.



\subsection{Tracking Controller}

A tracking MPC controller exploits the \textit{anytime} nature of BIT* to regularly query the current BIT* result and attempts to follow the collision-free obstacle-avoidance path while enforcing basic kinematic and velocity constraints. As our MPC is based on classical methods and is framed similarly to \cite{Zhou2022}, we do not provide specific implementation details in the interest of space.

\section{EXPERIMENTS}

\subsection{Performance Metrics}

In teach-and-repeat applications, it is critical that when avoiding obstacles we limit lateral path error to best exploit the prior knowledge on terrain assessment. It is also important to maintain a similar sensor viewing angle throughout the trajectory for localization purposes. With these characteristics in mind, we propose two metrics to compare the relative quality of the path solutions produced over the course of our experiments. We compute the Root Mean Square Error (RMSE) for both the translation and rotation (heading) components relative to the reference path over 15 m segments of obstacle interactions and average the result across all trials.


\subsection{Unit Testing of New Edge Cost}

We demonstrate the benefits of the lateral edge metric by testing BIT* on several representative obstacle-avoidance problems, both with and without the extensions proposed in Section III. For fair comparison, we implemented our own C++ version of BIT* in accordance with \cite{Gammell2015}, using parameters of 150 samples per batch and an RGG constant of 1.1. Our extended implementation, Lateral BIT*, uses the laterally weighted edge-cost metric \eqref{eq:eight} with $\alpha = 0.5$  and the rectangular approximation of the informed sampling region. 

\begin{figure}[b]
	\addtolength{\abovecaptionskip}{-4pt}
	\centering
	\includegraphics[scale=0.50]{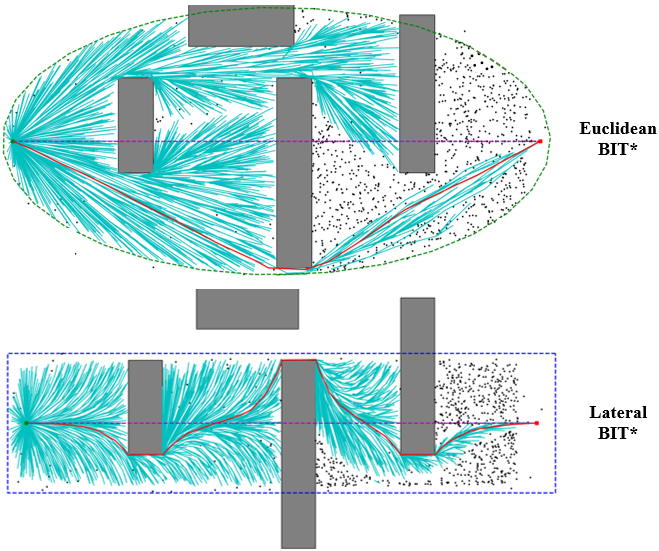}
	\caption{A comparison of the planning trees generated running BIT* with the Euclidean edge metric (top) and with the laterally weighted metric (bottom) in a representative test environment. Our edge metric encourages the final plan, shown in red, to avoid obstacles while remaining close to the magenta reference path on the $p$-axis.}
	\label{figure5}
\end{figure}

To study the influence of the new edge metric in isolation, we posed a series of ten simulated planning problems where the reference path is composed of a horizontal line 15 m in length on the $x$-axis, making the curvilinear and Euclidean path representations identical. We allow both versions of the algorithm to find converged path solutions connecting the starting pose to the goal and evaluate the solution both quantitatively and qualitatively. A representative example of the output paths and associated BIT* trees on one test problem is shown in Fig. 5.

In analyzing Fig. 5, we see some important differences in the exploration strategies of the two methods. Using our lateral edge-cost metric, BIT* tends to naturally generate smoothly curving solutions, spending the most time exploring paths near to the reference while balancing forward progress with reducing lateral error. In contrast, the standard BIT* algorithm settles on the direct shortest-distance solution. While efficient, in practice this path could be a higher-risk manoeuvre due to the additional localization and terrain-assessment uncertainty incurred when away from the reference path. We calculate the RMSE metrics for both implementations and summarize the results in Table 1.

\begin{table}[t]
	\caption{Planner Path Error Analysis For Simulations}
	\label{table_example}
	\begin{center}
		\scalebox{0.8}{
			\begin{tabular}{|c||c|c||c|c|}
				\hline
				\textbf{Exp. \#}	& \specialcell[c]{\textbf{Trans.}\\\textbf{RMSE [m]}\\\textbf{(Lateral BIT*)}} & \specialcell[c]{\textbf{Trans.}\\\textbf{RMSE [m]}\\\textbf{(Euclid BIT*)}} & \specialcell[c]{\textbf{Rot.}\\\textbf{RMSE [rad]}\\\textbf{(Lateral BIT*)}} & \specialcell[c]{\textbf{Rot.}\\\textbf{RMSE [rad]}\\\textbf{(Euclid BIT*)}}\\
				\hline
				1. & 0.085 & 0.268 & 0.513 & 0.622\\
				\hline
				2. & 0.101 & 0.221 & 0.573 & 0.516\\
				\hline
				3. & 0.0975 & 0.247 & 0.527 & 0.605\\
				\hline
				4. & 0.099 & 0.310 & 0.436 & 0.555\\
				\hline
				5. & 0.071 & 0.226 & 0.415 & 0.442\\
				\hline
				6. & 0.112 & 0.251 & 0.721 & 0.571\\
				\hline
				7. & 0.108 & 0.267 & 0.507 & 0.619\\
				\hline
				8. & 0.095 & 0.249 & 0.536 & 0.543\\
				\hline
				9. & 0.089 & 0.270 & 0.484 & 0.761\\
				\hline
				10. & 0.123 & 0.237 & 0.623 & 0.504\\
				\hline
				\textbf{\textbf{Mean:}} & \textbf{0.098} & \textbf{0.254} & \textbf{0.534} & \textbf{0.574}\\
				\hline
				
		\end{tabular}}
	\end{center}
	\vspace*{-0.35cm}
\end{table}

As we would expect, the use of our laterally weighted edge metric considerably reduces the average lateral error with respect to the reference path from 0.254 m to 0.098 m. We also see a small improvement on the heading error, likely due to the fact the lateral planner tends to spend more time exactly following the reference path. In terms of computation time, Lateral BIT* was able to find initial solutions to the problems in 40 ms or less, while on average exceeding 97\% solution convergence in just 0.2 seconds.

\begin{figure}[b]
	\centering
	\addtolength{\abovecaptionskip}{-4pt}
	\includegraphics[scale=0.45]{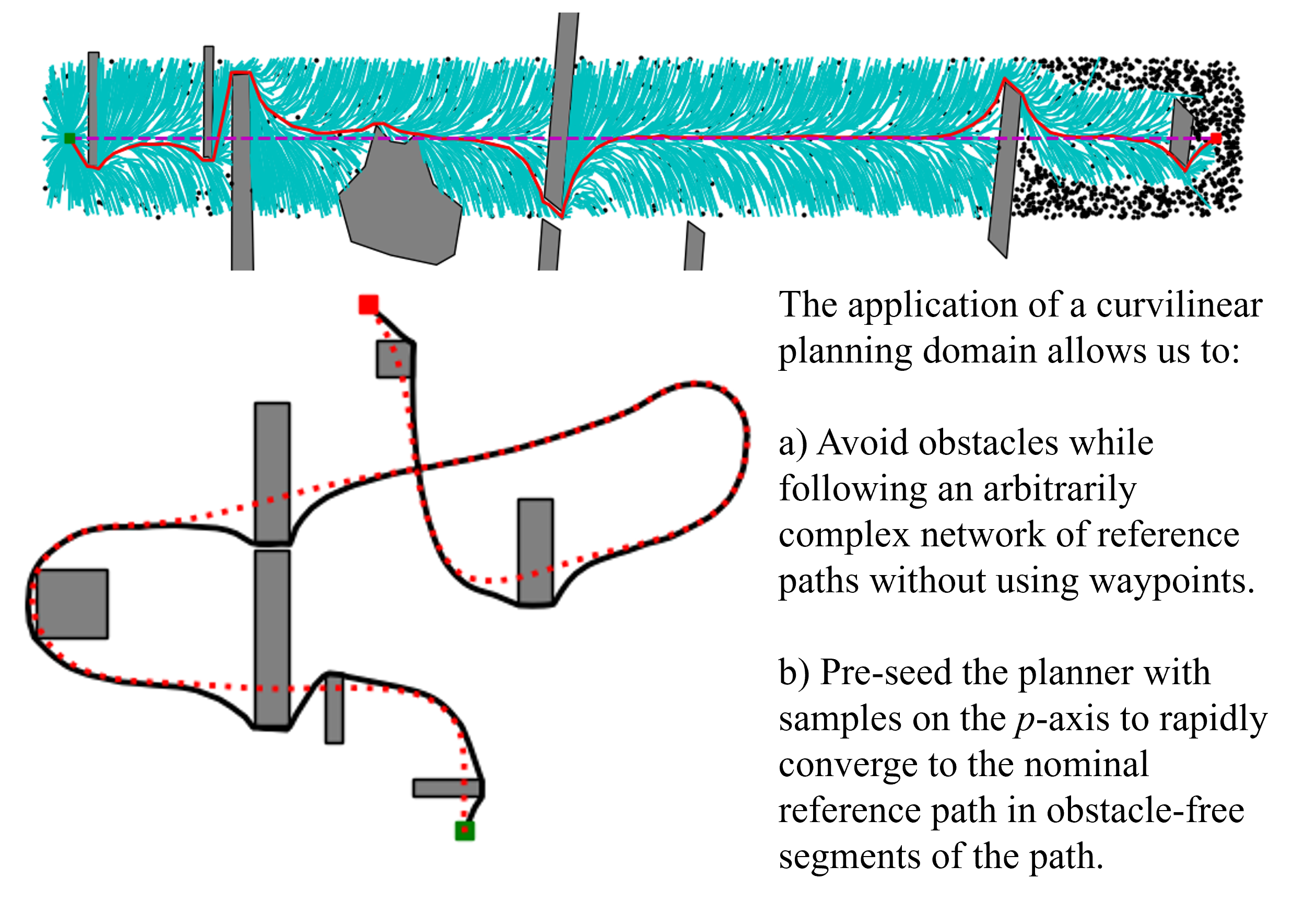}
	\caption{Lateral BIT* planning in a curvilinear representation of the space (top) to find a global obstacle avoidance path solution in Euclidean space (bottom) along a complex reference path with many obstacles.}
	\label{figure6}
\end{figure}

While it is easy to see the desirable output path properties the lateral edge metric encourages using straight-line reference paths, we can further exploit the curvilinear coordinate space to produce similarly smooth plans on more intricate reference paths. In Fig. 6, we initialize the planner on a complex path taken from real teach data that includes a variety of sharp curves, a path crossing, and several difficult obstacles. Despite the challenges, our planner is able to converge to a desirable solution using only a single goal and no intermediate waypoints. As Euclidean BIT* tends to generate solutions that traverse singularity regions in curvilinear space (this subsequently produces discontinuous Euclidean paths), we are not able to directly provide a comparison to this result.

\subsection{Field Trials}
\setlength{\footnotesep}{0pc}
Our final experiment evaluates the viability of the entire system in real-world settings. We integrated our obstacle-avoidance architecture into the VT\&R3 codebase\footnote{Code at https://github.com/utiasASRL/vtr3} and performed long-term autonomy tests in three different environments at the University of Toronto Institute for Aerospace Studies campus. Experiments were performed on a Clearpath Robotics Grizzly differential-drive robot equipped with a single Waymo Honeycomb LiDAR operating at 5Hz.


\begin{figure}[b]
	\centering
	\includegraphics[scale=0.34]{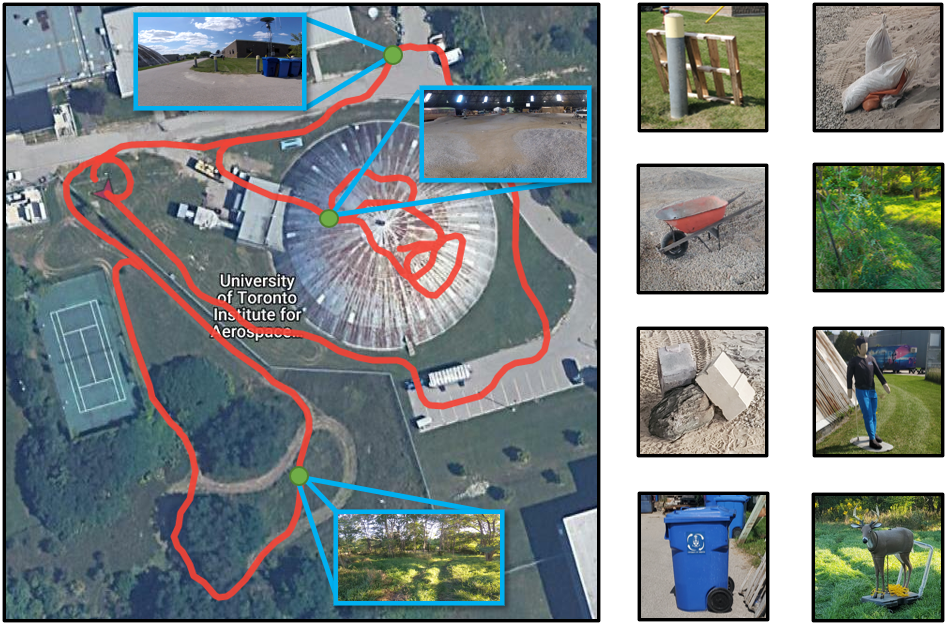}
	\caption{A satellite view of the network of paths used for the reference paths in the field trials. After teaching, we populated the reference paths with a diverse set of obstacles to avoid, ranging from natural elements such as rocks and fallen tree branches, to human analogs.}
	\label{figure7}
\end{figure}

The robot was manually driven to teach an obstacle-free network of paths of approximately 1.1 km, as shown in Fig. 7. A variety of obstacles were scattered throughout the network of paths to obstruct the nominal routes. Across three separate navigation sessions (each one beginning from a unique test environment), the robot repeated the network of paths a total of five times at 0.85 m/s, successfully avoiding obstacles and completing the routes with a 100\% autonomy rate. Some examples of typical obstacle interactions are shown in Fig. 8.

\begin{figure}[t]
	\centering
	\includegraphics[scale=0.36]{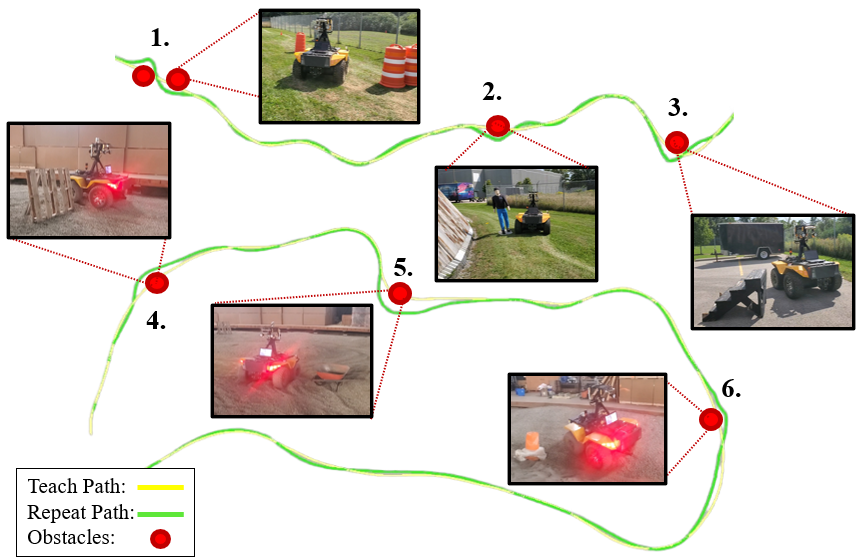}
	\caption{A visualization of the actual robot path (green) relative to the reference path (yellow) is provided over two short segments of the path network to offer a sense of the systems navigation tendencies. Typical obstacle interaction sites are highlighted in red and are labelled with the associated imagery from live experiments.}
	\label{figure8}
\end{figure}

We show the actual robot path (green) compared to the taught reference path (yellow) over two representative segments of the path network to give a sense of the true behaviour of our entire system. Similar to the simulated output path of Fig. 6, the planner tends to hug the reference path and then smoothly transitions around obstacles. Over obstacle-free stretches of the path, the planner returns to the nominal reference path. We can quantify the performance of the planner across these interactions by first calculating the RMSE for an obstacle-free repeat of the environment to get a baseline for the path-tracking performance. We then recompute the RMSE metrics for the obstacle-avoidance interactions to see the relative change. An obstacle interaction is defined as the length of path extending 7.5 m on either side of the obstacle for the purposes of RMSE calculation and we provide the obstacle's diameter for context. A subset of the obstacle interaction results are provided in Table 2. 

\begin{table}[tbhp]
	\caption{Obstacle-Free vs. Obstacles Error Analysis For Experiments}
	\label{table_example}
	\begin{center}
		\scalebox{0.8}{
			\begin{tabular}{|c|c||c|c||c|c|}
				\hline
				\specialcell[c]{\textbf{Obs.}\\\textbf{\#}}&
				\specialcell[c]{\textbf{Obs.}\\\textbf{Diameter}\\\textbf{[m]}}& \specialcell[c]{\textbf{Trans.}\\\textbf{RMSE [m]}\\\textbf{(obstacle-free)}} & \specialcell[c]{\textbf{Trans.}\\\textbf{RMSE [m]}\\\textbf{(obstacles)}} & \specialcell[c]{\textbf{Rot.}\\\textbf{RMSE [rad]}\\\textbf{(obstacle-free)}} & \specialcell[c]{\textbf{Rot.}\\\textbf{RMSE [rad]}\\\textbf{(obstacles)}}\\
				\hline
				1. & 1.8 & 0.031 & 0.121 & 0.052 & 0.352\\
				\hline
				2. & 0.6 & 0.046 & 0.077 & 0.064 & 0.219\\
				\hline
				3. & 1.0 & 0.073 & 0.123 & 0.092 & 0.394\\
				\hline
				4. & 1.4 & 0.042 & 0.112 & 0.057 & 0.258\\
				\hline
				5. & 1.1 & 0.056 & 0.110 & 0.079 & 0.324\\
				\hline
				6. & 0.4 & 0.075 & 0.095 & 0.082 & 0.149\\
				\hline
				
		\end{tabular}}
	\end{center}
	\vspace*{-0.35cm}
\end{table}

The data tend to agree with our qualitative observations that the planner does well to avoid excessive translation and rotation errors relative to the size of the obstacles during avoidance manouvres, with no localization failures caused as a result. Another interesting measure is the maximum lateral deviation per obstacle interaction. On average, our planner deviated from the path laterally a maximum of $r + 0.346$ m with a standard deviation of $\pm$0.074 m, where $r$ is the lateral extent of the obstacle geometry obstructing the reference path. The magnitude is directly related to our obstacle inflation tuning parameter of 0.3 m, but more importantly we observe that our system is relatively consistent when handling obstacles of varying sizes and geometries.

\section{CONCLUSIONS AND FUTURE WORK}
In this work, we presented a local obstacle-avoidance architecture designed for path-following applications such as teach and repeat. By modifying a sample-based motion planner to use a laterally weighted edge-cost metric combined with a curvilinear planning space, we show how natural path-following can be achieved that exploits prior knowledge of the terrain to avoid obstacles. Future work seeks to couple the MPC to the planner with dynamic-corridor state constraints as opposed to direct tracking. We believe such techniques will allow us to form stronger guarantees on collision avoidance as well as better handle moving obstacles. While early results are promising, we plan to continue validating our approach with additional experiments on multiple robots to continue pushing the boundaries of long-term autonomy.








\section*{ACKNOWLEDGMENT}

This research is supported by the Natural Sciences and Engineering Research Council of Canada (NSERC) and the Vector Scholarship in Artificial Intelligence.


\bibliographystyle{IEEEtran}
\bibliography{IEEEabrv,bibliography_file}

\end{document}